\documentclass[sigconf]{acmart}

\AtBeginDocument{%
  \providecommand\BibTeX{{%
    \normalfont B\kern-0.5em{\scshape i\kern-0.25em b}\kern-0.8em\TeX}}}

\setcopyright{acmcopyright}
\copyrightyear{2023}
\acmYear{2023}
\acmDOI{XXXXXXX.XXXXXXX}

\usepackage{subfig}
\usepackage{bm}
\usepackage{amsthm}
\usepackage{algorithm}
\usepackage{algorithmic}
\usepackage{pifont}
\usepackage{multirow}
\usepackage{array}
\usepackage{booktabs}

\usepackage{bbm}
\usepackage{colortbl}  
\usepackage{xcolor}
\usepackage{array} 

\renewcommand{\paragraph}[1]{\vspace{1ex}\noindent\textbf{#1}.}
\newcommand{\sysname}{SGMPT}
\begin{document}

\title{Structure Guided Multi-modal Pre-trained Transformer for Knowledge Graph Reasoning}

\author{Ke Liang}
\affiliation{%
  \institution{National University of Defense Technology}
  \city{Changsha}
  \state{Hunan}
  \country{China}
}

\author{Sihang Zhou}
\affiliation{%
  \institution{National University of Defense Technology}
  \city{Changsha}
  \state{Hunan}
  \country{China}
}

\author{Yue Liu}
\affiliation{%
  \institution{National University of Defense Technology}
  \city{Changsha}
  \state{Hunan}
  \country{China}
}

\author{Lingyuan Meng}
\affiliation{%
  \institution{National University of Defense Technology}
  \city{Changsha}
  \state{Hunan}
  \country{China}
}

\author{Meng Liu}
\affiliation{%
  \institution{National University of Defense Technology}
  \city{Changsha}
  \state{Hunan}
  \country{China}
}

\author{Xinwang Liu}
\authornotemark[1]
\affiliation{%
  \institution{National University of Defense Technology}
  \city{Changsha}
  \state{Hunan}
  \country{China}
}

\renewcommand{\shortauthors}{}

\begin{abstract}
Multimodal knowledge graphs (MKGs), which intuitively organize information in various modalities, can benefit multiple practical downstream tasks, such as recommendation systems, and visual question answering. However, most MKGs are still far from complete, which motivates the flourishing of MKG reasoning models. Recently, with the development of general artificial architectures, the pretrained transformer models have drawn increasing attention, especially for multimodal scenarios. However, the research of multimodal pretrained transformer (MPT) for knowledge graph reasoning (KGR) is still at an early stage. As the biggest difference between MKG and other multimodal data, the rich structural information underlying the MKG still cannot be fully leveraged in existing MPT models. Most of them only utilize the graph structure as a retrieval map for matching images and texts connected with the same entity. This manner hinders their reasoning performances. To this end, we propose the graph \underline{S}tructure \underline{G}uided \underline{M}ultimodal \underline{P}retrained \underline{T}ransformer for knowledge graph reasoning, termed \sysname{}. Specifically, the graph structure encoder is adopted for structural feature encoding. Then, a structure-guided fusion module with two different strategies, \textit{i.e.,} weighted summation and alignment constraint, is first designed to inject the structural information into both the textual and visual features. To the best of our knowledge, \sysname{} is the first MPT model for multimodal KGR, which mines the structural information underlying the knowledge graph. Extensive experiments on FB15k-237-IMG and WN18-IMG, demonstrate that our \sysname{} outperforms existing state-of-the-art models, and prove the effectiveness of the designed strategies.
\end{abstract}




\maketitle

\begin{figure}[t]
\centering
\includegraphics[width=0.48\textwidth]{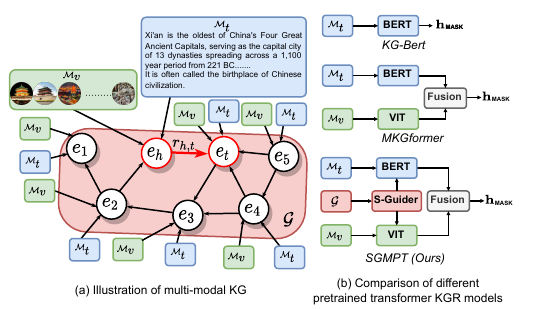}
\caption{Illustration of multimodal knowledge graph and existing pretrained transformer KGR models. $\mathcal{M}_t$, $\mathcal{M}_v$, and $\mathcal{G}$ represent the textual information, visual information, and graph structural information, respectively. Besides, BERT \cite{BERT} and ViT \cite{VIT} are two commonly used transformers for encoding $\mathcal{M}_t$ and $\mathcal{M}_v$, and structure-guided fusion module (S-Guider) is a novel module designed in our \sysname{} to leverage $\mathcal{G}$. Note that $e$ and $r$ denote as entity and relation, separately.}
\label{Illu_IR}  
\end{figure}

\section{Introduction}
Multimodal knowledge graphs (MKGs), which intuitively organize information in various modalities, can benefit many practical downstream tasks, such as recommendation systems \cite{Recommander1, Recommander2}, information retrieval \cite{informationretrieval1, informationretrieval2}, and visual question answering \cite{VQA1}. Specifically, compared to traditional KGs, extra multimodal data are linked with entities in MKGs to provide more meaningful information, such as visual and textual attributes, which makes them more closely to the real world. However, existing multimodal KGs may suffer from even more severe incompleteness issues due to the insufficient accumulation of multimodal corpus. It compromises their usability and impairs their effectiveness. To address the limitations, more and more recent attempts \cite{DKRL, MKGformer, MM22MKGR, MMKGR, MoSE, TransAE} for knowledge graph reasoning (KGR) are taken in multimodal scenarios, and our work also lays in this scope.

The incompleteness issues, as a common nature among different types of KGs, have been widely studied these years \cite{AKGR}. According to previous attempts for KGR, we can easily draw a conclusion that the performance of the KGR model highly depends on whether the information in KGs is fully leveraged or not. For example, the conventional KGR models \cite{NCRL, COMPGCN, GraIL, CURL, liang2023knowledge, liang2023abslearn, liang2023message, meng2023sarf, KG1_Luo, KG2_Luo} focus on mining the structural information in static KGs. However, an additional effective mechanism for temporal information is important for these models to achieve better reasoning in temporal KGs. It is similar to multimodal knowledge graph reasoning (MKGR), which requires specific mechanisms to mine the multimodal information since previous unimodal KGR models can also not scale well to multimodal scenarios. Following this trace, different mechanisms are designed based on the unimodal KGR models to fuse the heterogeneous multimodal features of each entity into a unified representation, which builds up various multimodal KGR models. For example, IKRL \cite{IKRL} and MTRL \cite{MTRL} are both developed based on TransE \cite{TransE} but with different fusion strategies. The former model adopts the attention mechanism to integrate the visual information, while the fusion module in the latter one offers three different fusion options. However, with the development of general artificial intelligence (AGI), the pretrained transformer architecture has drawn increasing attention as a more general and powerful paradigm, especially for multimodal scenarios. Inspired by their success in other fields, lots of pretrained transformer models for KGR have come out these years, such as KG-BERT \cite{KG-BERT}, KGformer \cite{kgformer}, and etc. However, the research on developing an effective multimodal pretrained transformer (MPT) for KGR is still at an early stage, which leaves us a huge gap to explore.

Among the MPT models for KGR, MKGformer  \cite{MKGformer} is the most representative model with the best reasoning capacity. Although achieving promising performance for MKGR, it still fails to fully leverage the structural information underlying the knowledge graph, which hinders their reasoning capacity. Unlike other multimodal data, MKG usually contains three types of information, \textit{i.e.,} textual information $\mathcal{M}_t$ (\textit{e.g.,} text description), visual information $\mathcal{M}_v$ (\textit{e.g.,} images), and graph structural information $\mathcal{G}$ as shown in Fig. \ref{Illu_IR} (a). MPT models for other multimodal scenarios only take consideration of the first two modalities, which is also how MKGformer does for MKGR task as shown in Fig. \ref{Illu_IR} (b). More specifically, the graph structure of MKG is only utilized as a retrieval map for matching images and text descriptions corresponding to the same entity. In this manner, the rich structural information underlying the graph structure $\mathcal{G}$ is ignored, such as the relational information between different entities and the topological information within the graph structure. This structural information will definitely benefit the expressive ability of the models, which has been proven in those multimodal non-transformer KGR models \cite{MoSE, TransAE}. As it is currently ignored by MPT models for MKGR, all we need is to design an effective fusion mechanism to mine such structural information, which will endows the existing MPT models greater capacity for better reasoning performance.

Following this idea, we propose a novel graph \underline{S}tructure \underline{G}uided \underline{M}ultimodal \underline{P}retrained \underline{T}ransformer model for knowledge graph reasoning, termed \sysname{}, by designing a plug-and-play mechanism to leverage the structural information omitted by previous MPT models as shown in Fig. \ref{OVERRALL_FIGURE}. More specifically, the graph structure encoder is adopted for structure feature encoding. Then, a structure-guided fusion module with two different strategies, \textit{i.e.,} weighted summation and alignment constraint, is first designed to fuse the structure information in both the textual and visual features. Concretely, (1) weighted summation directly adds the generated structural embedding with the textual and visual embeddings in the segment for the entity. Besides, (2) the alignment constraint adopts the alignment loss, \textit{i.e.,} MSE loss \cite{ermolov2021whitening, BRGL}, to guide the learning procedure by refining original textual and visual embeddings according to structural information. Moreover, the above strategies can be adopted both individually and composedly. To the best of our knowledge, \sysname{} is the first multimodal pretrained transformer for KGR, which tries to mine the structural information underlying the knowledge graph. In addition, extensive experiments are carried out on two benchmark datasets to demonstrate the capacity of \sysname{} from four aspects, \textit{i.e.,} superiority, effectiveness, efficiency, and sensitivity. The main contributions are summarized below:
\begin{itemize}
    \item We propose a novel and simple {g}raph {s}tructure {g}uided multimodal {p}retrained {t}ransformer model for knowledge graph reasoning, termed \sysname{}, by effectively making use of the knowledge graph structural information.

    \item We adopt the structural encoder and design a plug-and-play mechanism, \textit{i.e.,} structure-guided fusion module, is proposed to complement the omitted graph structural information for MPT models for knowledge graph reasoning. Specifically, the structure-guided fusion module is composed of two different strategies, \textit{i.e.,} weighted summation and alignment constraint, which can be adopted both individually and composedly.
    
    \item Extensive experiments on FB15k-237-IMG and WN18-IMG datasets demonstrate the capacity of \sysname{} from four aspects, \textit{i.e.,} superiority, effectiveness, efficiency, and sensitivity, and also proves the effectiveness of the structural information. 
\end{itemize}

\section{RELATED WORK}\label{rw}
\subsection{Multimodal Knowledge Graph Reasoning}
Multimodal knowledge graph reasoning (MKGR) aims to infer the potential missing facts in multimodal knowledge graph, which can be roughly divided into two types, \textit{i.e.,} non-transformer models and transformer models, according to the model architectures.

\subsubsection{Non-Transformer Multimodal KGR Models}
Most of the MKGR models are developed based on non-transformer architectures. Different mechanisms are designed to encode the extra modal information by extending the original unimodal KGR models, such as TransE \cite{TransE}.
For example, IKRL \cite{IKRL} first adopts the attention based mechanism to integrate the visual information and the original structural information generated by the translation based KGR models. MTRL \cite{MTRL} offers three different strategies, \textit{i.e.,} simple summation, DeViSE \cite{DeViSE}, and Imagined \cite{Imagined} to integrate multimodal information. In addition, TransAE \cite{TransAE} utilized an auto-encoder to use them. Moreover, MoSE \cite{MoSE} exploits three ensemble inference techniques to combine the modality-split predictions by assessing modality importance. Recently, RSME \cite{RSME} designed a forget gate with an MRP metric to select valuable images for multimodal KGR, which tries to avoid the influence caused by the noise from irrelevant images corresponding to entities. However, our model do not belong to this type.

\subsubsection{Transformer Multimodal KGR Models}
The transformer models originated form natural language processing \cite{BERT, GPT2, GPT3}, and quickly shifted the paradigm of image processing \cite{bao2022beit, VIT} from fully supervised learning to pretraining and fine-tuning. Due to their promising capabilities in multimodal scenarios, various general multimodal pretrained transformer (MPT) models \cite{Fashionbert, VL-Bert, visualbert, Unicoder-vl, Uniter, Lxmert, Vilbert} have been proposed these years. However, the target optimization objects of the above multimodal pretrained models are less relevant to knowledge graph reasoning (KGR) tasks. Due to the variance between the multimodal knowledge graph (MKG) and other multimodal data, directly applying the above general MPT models to multimodal knowledge graph reasoning (MKGR) may not lead to good reasoning \cite{MKGformer}. 

Meanwhile, pretrained transformer models \cite{HittER, MKGformer, TET, koncel2019text, li2022kpgt, kgformer, sun2022graph} for KGR are also springing up, such as KG-BERT \cite{KG-BERT}, which is the first pretrained contextual language model for the KGR task, etc. However, the research on developing an effective multimodal pretrained transformer (MPT) for KGR is still at an early stage. Among them, MKGformer \cite{MKGformer} is the most representative attempt with promising reasoning capacity, which leaves us huge space to explore better MPT models for KGR. But MKGformer \cite{MKGformer} also ignores the key difference of the characteristic between MKG and other multimodal data. Specifically, unlike other multimodal data, MKG usually contains three types of information, \textit{i.e.,} textual information $\mathcal{M}_t$ (\textit{e.g.,} text description), visual information $\mathcal{M}_v$ (\textit{e.g.,} images), and graph structural information $\mathcal{G}$ as shown in Fig. \ref{Illu_IR} (a). MPT models for other multimodal scenarios only take consideration of the first two modalities, which is also how MKGformer does for MKGR task as shown in Fig. \ref{Illu_IR} (b). In other words, the graph structure of MKG is only utilized as a retrieval map for matching images and text descriptions corresponding to the same entity. In this manner, the rich structural information underlying the knowledge graph $\mathcal{G}$ is ignored, such as the relational information between different entities and the topological information within the graph structure. This structural information will definitely benefit the expressive ability of the models, which has been proven in those multimodal non-transformer KGR models \cite{MoSE, TransAE}. As it is currently ignored by MPT models for MKGR, all we need is to design an effective fusion mechanism to mine such structural information for multimodal transformer KGR models. To this end, our work endows the existing MPT models greater capacity for those omitted structural information to achieve better reasoning.

\subsection{Multimodal Fusion Strategy}
Information fusion aims to integrate information from different modalities to contribute to the specific downstream tasks \cite{MultimodalMLSurvey}, which is usually treated as one of the important step for multimodal and multisource tasks. There are generally two ways to fuse the features, \textit{i.e.,} (1) combining every single modal representation in its own feature space, such as summation, average pooling \cite{MMKGSURVEY151, MMKGSURVEY39, MMKGSURVEY40}, and (2) learning the unified representations by projecting different modal representations into the same latent space based on a well-designed objective function \cite{MMKGSURVEY145, MTRL, IKRL}. Inspired by them, our \sysname{} is the first MPT model to fuse the structural information with the original textual and visual information by introducing a plug-and-play structure-guided fusion module. Two strategies are designed in the structure-guided fusion module, including weighted summation and alignment constraint. Concretely, weighted summation belongs to the first type, while the alignment constraint belongs to the second type. Both above two types of strategies can be adopted individually and composedly. More details about the fusion strategies are illustrated in the next section.

\section{Method}
\begin{figure*}[t]
\centering
\includegraphics[width=\textwidth]{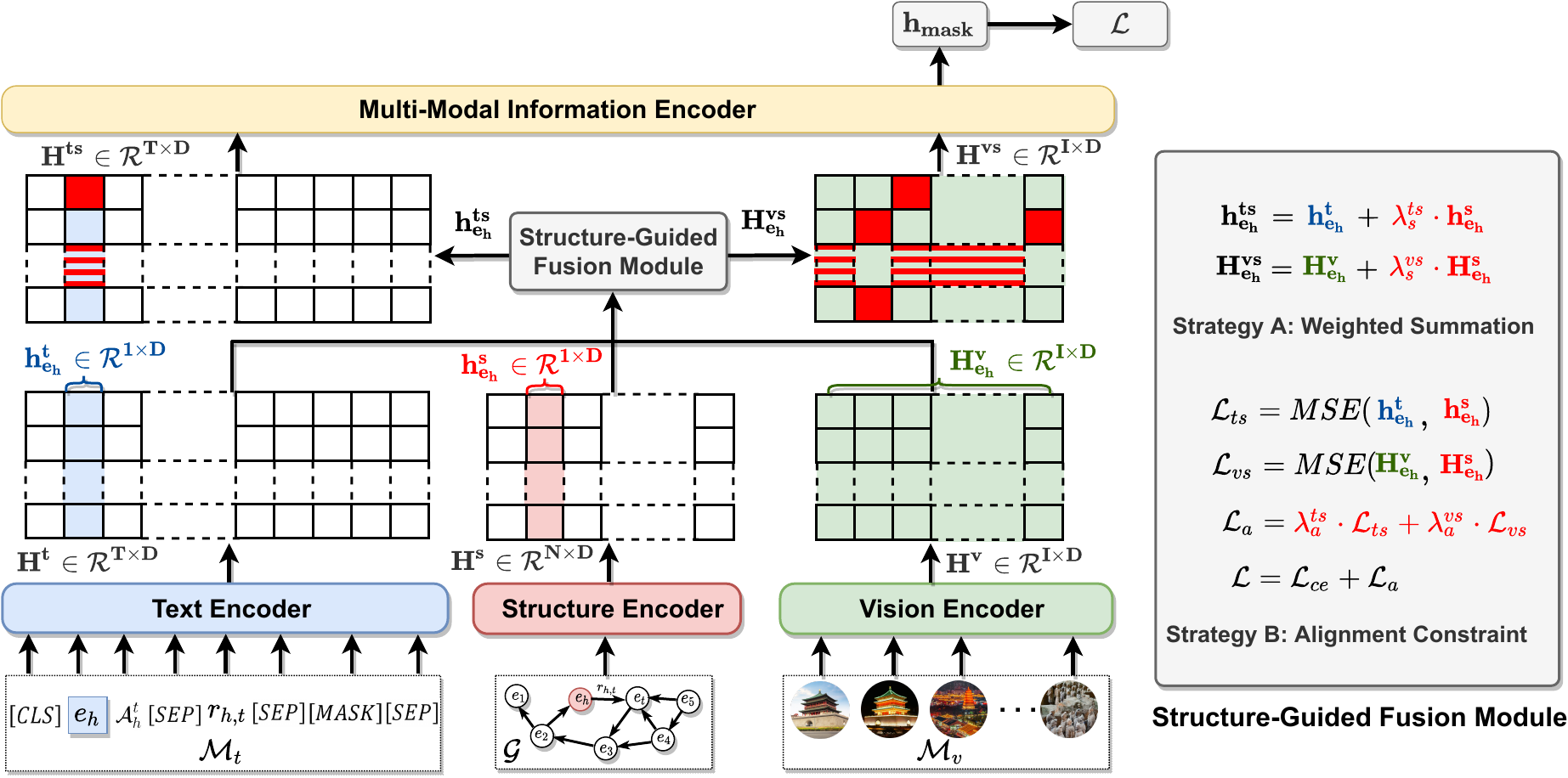}
\caption{The framework of the proposed graph \underline{S}tructure \underline{G}uided \underline{M}ultimodal \underline{P}retrained \underline{T}ransformer model for knowledge graph reasoning, termed \sysname{}. The structure encoder and structure-guided fusion module are proposed to complement the omitted graph structural information for MPT models for knowledge graph reasoning. Precisely, the structure-guided fusion module is a plug-and-play mechanism, which contains two different strategies, \textit{i.e.,} weighted summation and alignment constraint. The blue, green, and red colors represent the textual, visual, and structural features, respectively. Note that more details of the fusion strategies can be found in Fig. \ref{EXAMPLE} and Tab. \ref{notation} presents the descriptions of notations.}
\label{OVERRALL_FIGURE}  
\end{figure*}

In this section, we will introduce the details of the proposed graph structure guided multimodal pretrained transformer model, termed \sysname{}, from three aspects, \textit{i.e.,} preliminary, multimodal pretrained transformer backbone, and guidance of structure information. The overall framework of our \sysname{} is shown in Fig. \ref{OVERRALL_FIGURE}, and the pseudocode is shown in Algorithm \ref{ALGORITHM}.
\begin{table}[t]
\caption{The Summary of important notations}
\vspace{-1em}
\fontsize{23}{35}\selectfont 
\label{notation}
\resizebox{\linewidth}{!}{
\begin{tabular}{cc}  
\hline 
Notations & Descriptions \\  
\hline   
$\emph{MKG}=(\mathcal{E}, \mathcal{R}, \mathcal{G}, \mathcal{A}_{\mathcal{M}})$ & multimodal knowledge graph  \\  
$\mathcal{E}$ & entity set  \\  
$\mathcal{R}$ & relation set  \\
${\mathcal{G}}$ & graph with fact triplets (edges)  \\  
$\mathcal{A}_{\mathcal{M}}$ & multimodal attribute set  \\  
$\mathcal{M}_t,\mathcal{M}_v$ & textual and visual modality \\ 
$A^t_{h}=\{w_1, w_2,\cdots,w_n\}$ & textual attribute, $w_i$ represents the $i^{th}$ word  \\ 
$A^v_{h}=\{I_1, I_2,\cdots, I_m\}$ & visual attribute, $I_i$ represents the $i^{th}$ image\\
$\textbf{H},\textbf{h}$ & feature matrix, feature vector \\ 
$\textbf{H}^t\in\mathcal{R}^{T \times D}$& textual feature matrix, $T$ is the number of tokens\\ 
$\textbf{H}^v\in\mathcal{R}^{I \times D}$& visual feature matrix, $I$ is the number of images \\   
$\textbf{H}^s\in\mathcal{R}^{N \times D}$& structural feature matrix, $N$ is the number of entities\\  
$\textbf{H}^{ts},\textbf{H}^{vs}$& text-structure and vision-structure feature matrix\\  
$\textbf{h}_{mask}$& representation of {{[MASK]}}  \\ 
$\mathcal{L}$& total loss  \\  
$\mathcal{L}_{ts}$, $\mathcal{L}_{vs}$& text-structure and vision-structure alignment loss  \\ 
$\mathcal{L}_a$, $\mathcal{L}_{ce}$& alignment loss, cross-entropy loss  \\ 
$\lambda_{s}^{ts}$, $\lambda_{s}^{vs}$& hyperparameter for weighted summation  \\  
$\lambda_{a}^{ts}$, $\lambda_{a}^{vs}$& hyperparameter for alignment constraint \\ 
\hline 
\end{tabular}}
\end{table}

\subsection{Preliminary}
The multimodal knowledge graph is defined as a directed graph \emph{MKG} $=(\mathcal{E}, \mathcal{R}, \mathcal{G}, \mathcal{A}_{\mathcal{M}})$, where $\mathcal{E}$ and $\mathcal{R}$ represent the entity set and the relation set respectively, and $\mathcal{G}=\{(e_h,r_{h,t},e_t)\ | \ e_h,e_t \in \mathcal{E}, r_{h,t} \in \mathcal{R}\}$ is the set of fact triplets. $\mathcal{A}_{\mathcal{M}}$ represents the set of multimodal attributes corresponding to each entity, which contains two types of modalities, \textit{i.e.,} text descriptions ($\mathcal{M}_t$) and image descriptions ($\mathcal{M}_v$). Given the missing facts $(e_h, r_{h,t}, ?)$, the main goal of knowledge graph reasoning (MKGR) is to infer the entity $e_t$ based on the \emph{MKG}. Similar to previous multimodal pretrained transformer (MPT) KGR models \cite{MKGformer}, the MKGR tasks in this paper can be divided into steps, including: (1) \textbf{Pretaining}: image-text incorporated entity representation learning, and (2) \textbf{Finetuning}: relation reasoning over multimodal entity representations. As shown in Eq. (1) and Eq. (2), the $[CLS]$ and $[SEP]$ are two separation symbols, and $[MASK]$ are the prediction symbol. Notably, the multimodal attributes, \textit{i.e.,} visual and structural features, are encoded and integrated into the textual feature, and both pretraining and finetuning tasks are still reformulated as masked language modeling (MLM) tasks based on the textual features.

\paragraph{Pretraining} The pretraining procedure aims to match the multimodal attributes with the masked corresponding entity $e_i$ as follows.
\begin{align}
{T(e_i)}={[CLS]\ A^t_{i}\ is\ the\ description\ of\ [MASK][SEP]}
\end{align}
    
\paragraph{Finetuning} For reasoning task $T(e_{h},r_{h,t},?)$, the main goal of the finetuning procedure is to predict the masked target entity $e_t$ as follows.
\begin{align}
{T(e_h,r_{h,t},?)}={[CLS]\ e_{h}\ A^t_{h}\ [SEP]\ r_{h,t}\ [SEP][MASK][SEP]}
\end{align}

\subsection{Multimodal Pretrained Transformer Backbone}
The MKGformer \cite{MKGformer}, which is the most representative MPT models for KGR, is selected as our MPT backbone. It consists of three different encoders, \textit{i.e.,} text encoder, vision encoder, and multimodal information encoder. Specifically, the number of layers in the text encoder, vision encoder, and multimodal information encoder are $L_t$, $L_v$, and $L_m$, respectively, where $L_{BERT}=L_t + L_m$ and $L_{ViT}=L_v + L_m$. We briefly introduce the important components of the backbone model as shown below, and more details can be found in \cite{MKGformer}.

\paragraph{Text Encoder}
Text encoder $f_{t}(\cdot)$ is composed of the first $L_t$ layers of BERT \cite{BERT}, which aims to capture basic syntactic and lexical information. It takes tokens in the text descriptions $\mathcal{M}_t$ as input, and outputs the textual feature $\textbf{H}^{t}$. 
\begin{align}
\textbf{H}^{t}=f_{t}(\mathcal{M}_t)
\end{align}

\paragraph{Vision Encoder}
Vision encoder $f_{v}(\cdot)$ is composed of the first $L_v$ layers of ViT \cite{VIT}, which aims to capture basic visual features from the patched images. It takes images $\mathcal{M}_v$ as input, and outputs the visual feature $\textbf{H}^{v}$.
\begin{align}
\textbf{H}^{v}=f_{v}(\mathcal{M}_v)
\end{align}

\paragraph{Multimodal Information Encoder}
Following \cite{MKGformer}, multimodal information encoder $f_m(\cdot)$ aims to model the multimodal features of the entity across the last $L_m$ layers of ViT and BERT with multi-level fusion. It takes learned representations from previous encoders as input, and outputs the multimodal representations for inference.

\subsection{Guidance of Structure Information}
To guide the learning procedure for the multimodal pretrained transformer (MPT) model for KGR with the structural information, two novel modules are proposed, \textit{i.e.,} structure encoder and structure-guided fusion module. The structure encoder aims to encode the structural information into the feature vector, and the structure-guided fusion module tends to fuse the omitted structural information into the existing MPT models for KGR.

\subsubsection{Structure Encoder}
The structure encoder takes the static KG $\mathcal{G}$ as input without multimodal attributes and outputs the structural representation $\textbf{H}^{s}\in\mathcal{R}^{N \times D}$ for fusion, where $N$ represents the number of the entities in KG. Different knowledge graph embedding models contribute to the candidate structure encoder $g(\cdot)$, such as HAKE \cite{HAKE}, ComPGCN \cite{COMPGCN}, Nodepiece \cite{galkin2022nodepiece}, etc. 
\begin{align}
\textbf{H}^{s}=g(\mathcal{G})
\end{align}This paper selects one of the state-of-the-art models, HAKE \cite{HAKE}, as the structure encoder to generate the structural features for most of the experiments. In addition, we also present the influence of different structure encoders in Sec. 4.3.2.

\subsubsection{Structure-Guided Fusion Module}
The structure-guided fusion module is designed to fuse the structural information. Two simple yet effective strategies are designed in the structure guide, \textit{i.e.,} weighted summation, and alignment constraint, which can be adopted both individually and composedly (See Fig. \ref{OVERRALL_FIGURE}). The details of each strategy will be illustrated as follows.
 \begin{figure}[t]
\centering
\includegraphics[width=0.49\textwidth]{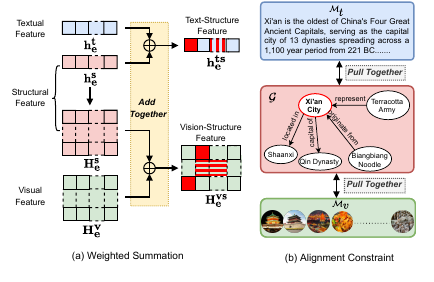}
\caption{The illustration of the designed strategies in structure-guided fusion module with an example of the entity 'Xi'an City'. Tab. \ref{notation} presents the descriptions of notations.}
\label{EXAMPLE}  
\end{figure}

\paragraph{Weighted Summation}
Weighted summation combines the generated structural embedding with the segment in textual and visual embeddings corresponding to the entity in its own feature space, as shown in Fig. \ref{EXAMPLE} (a). The summation procedure can be divided into two parts, \textit{i.e.,} (1) text-structure feature generation and (2) vision-structure feature generation. 

More specifically, for the first part, we first extract the specific token representation vector $\bf{h^{s}_{e_{h}}}\in\mathcal{R}^{1 \times D}$ from the structural feature matrix $\bf{H^{s}}\in \mathcal{R}^{N \times D}$ for each entity. Then, the fused feature vector is generated according to Eq. (4). 
\begin{equation} 
\bf{h_{e_h}^{ts}}=\bf{h_{e_h}^{t}}+\lambda_{s}^{ts}\cdot\bf{h_{e_h}^{s}},
\end{equation}where $\lambda_{s}^{ts}$ is the weighted hyper-parameter for text-structure feature generation. After that, we replace the feature vector $\bf{h_{e_h}^{ts}}$ for $\bf{h^{t}_{e_{h}}}$ in the original textual feature matrix and generate the text-structure feature matrix $\bf{H^{ts}}$. 
\begin{align}
\mathcal{L}=\mathcal{L}_{ce} + \mathcal{L}_{a}
\end{align}
\begin{algorithm}[!t]
\caption{Psedocode of \sysname{}}
\small
\label{ALGORITHM}
\textbf{Input}: A multimodal knowledge graph:
$\emph{MKG}=(\mathcal{E}, \mathcal{R}, \mathcal{G}, \mathcal{A}_{\mathcal{M}})$; The text encoder $f_t(\cdot)$; The vision encoder $f_v(\cdot)$; The multimodal information encoder $f_m(\cdot)$; The structure encoder $g(\cdot)$; Weights in the pretrained model $W$; Iteration number $t$; Hyperparameters $\lambda_{s}^{ts}$, $\lambda_{s}^{vs}$, $\lambda_{a}^{ts}$, and $\lambda_{a}^{vs}$.\\
\textbf{Output}: Weights in the fine-tuned model $W^*$.
\begin{algorithmic}[1]
\STATE Generate the structural representation $\bf{H^{s}}$ by Eq. (5);
\FOR{$i=1$ to $t$}
\FOR{$e$ in $\mathcal{E}$}
\STATE Extract the structural vector $\bf{h^{s}_e}$ for corresponding entity $e$ from $H^{s}$, and expand it to the matrix $\bf{H^{s}_e}$ by Eq. (7);
\STATE Generate the textual representation $\bf{H^{t}_e}$ for target entity by Eq. (3); 
\STATE Extract the token vector $\bf{h^{t}_e}$ from $\bf{H^{t}}$;
\STATE Generate the visual representation $\bf{H^{v}_e}$ for target entity by Eq. (4); 
\STATE Generate the text-structure vector $\bf{h^{ts}_{e_h}}$ by Eq. (6).
\STATE Obtain the text-structure feature $\bf{H^{ts}}$ by reassigning $\bf{h^{ts}_{e_h}}$ to the corresponding position in $\bf{H^{t}}$.
\STATE Generate the vision-structure feature $\bf{H^{vs}_{e_h}}$ (i.e., $\bf{H^{vs}}$) by Eq. (8).
\STATE Feed $\bf{H^{ts}}$ and $\bf{H^{vs}_{e_h}}$ into $f_m(\cdot)$ and get [MASK] embedding $\bf{h_\emph{MASK}}$;
\STATE Calculate alignment losses $\mathcal{L}_{ts}$ and $\mathcal{L}_{vs}$ by Eq. (9) and Eq. (10);
\STATE Calculate cross-entropy loss $\mathcal{L}_{ce}$ referred to \cite{MKGformer};
\STATE{Optimize the whole network $W$ by minimizing $\mathcal{L}$ in Eq. (12)};
\ENDFOR
\ENDFOR
\end{algorithmic}
\end{algorithm}

As for the second part, we first expand structural feature vector $\bf{h^{s}_{e_{h}}}\in\mathcal{R}^{1 \times D}$ to the structural feature matrix $\bf{H^{s}_{e_{h}}}\in\mathcal{R}^{I \times D}$ based on the Eq. (8) below.
\begin{equation} 
\bf{H^{s}_{e_{h}}}=\bf{h_{e_h}^{s}}\bf{{1}_{I}^{T}},
\end{equation}where $\bf{H^{s}_{e_{h}}}$ has the same dimension as the visual feature matrix $\bf{H^{v}_{e_{h}}}$ for each entity and $\bf{{1}_{I}}$ is the row vector with all elements set to 1. 

Then, the fused vision-structure feature matrix $\bf{H^{vs}}$ is generated according to Eq. (8).
\begin{equation} 
\bf{H_{e_h}^{vs}}=\bf{H_{e_h}^{v}}+\lambda_{s}^{vs}\cdot\bf{H^{s}_{e_{h}}},
\end{equation}where $\lambda_{s}^{vs}$ is another weighted hyper-parameter for vision-structure feature matrix generation.

\paragraph{Alignment Constraint}
Alignment constraint assists in learning unified representations by aligning the textual and visual feature representations to the structural feature representation using a self-supervised alignment loss, \textit{i.e.,} MSE loss \cite{ermolov2021whitening, BRGL}. The loss helps pull together different embeddings from different modalities. Similar to the weighted summation, the alignment constraint can also be divided into two parts, \textit{i.e.,} (1) text-structure feature alignment and (2) vision-structure feature alignment. 

Concretely, two types of loss functions are adopted, including text-structure alignment loss $\mathcal{L}_{ts}$ and vision-structure alignment loss $\mathcal{L}_{vs}$.
\begin{equation}
\begin{aligned}
\mathcal{L}_{ts} &= {{MSE}(\bf{h_{e_h}^{t}}, \bf{h_{e_h}^{s}})}\\ &= {\left\| \frac{\bf{h_{e_h}^{t}}}{\bf{\|h_{e_h}^{t}\|}_2} - \frac{\bf{h_{e_h}^{s}}}{\bf{\|h_{e_h}^{s}\|}_2} \right\|}_2^2 \\&= 2-2 \cdot {\frac{\left<\bf{h_{e_h}^{t}, h_{e_h}^{s}}\right>}{\|\bf{h_{e_h}^{t}}\|_2 \cdot \|\bf{h_{e_h}^{s}}\|_2}},
\end{aligned}
\label{positive_loss2}
\end{equation}
\begin{equation}
\begin{aligned}
\mathcal{L}_{vs} &= {{MSE}(\bf{H}_{e_h}^{v}, {\bf{H_{e_h}^{s}}})}\\ &= {\left\| \frac{\bf{H}_{e_h}^{v}}{\bf{\|\bf{H}_{e_h}^{v}\|}_2} - \frac{{\bf{H_{e_h}^{s}}}}{\bf{\|{\bf{H_{e_h}^{s}}}\|}_2} \right\|}_2^2 \\&= 2-2 \cdot \frac{\left<\bf{\bf{H}_{e_h}^{v}, {\bf{H_{e_h}^{s}}}}\right>}{{\|\bf{H}_{e_h}^{v}\|}_2 \cdot \|{\bf{H_{e_h}^{s}}}\|_2},
\end{aligned}
\label{positive_loss2}
\end{equation}where $\|\cdot\|$ denotes the 2-norm. Our network is optimized by minimizing the contrastive loss and $\bf{H_{e_{h}}^{s}}$ is generated by Eq. (8). Moreover, the total loss $\mathcal{L}_{a}$ for alignment is shown below:
\begin{equation} 
\mathcal{L}_{a} = \lambda_{a}^{ts}\cdot\mathcal{L}_{ts} + \lambda_{a}^{vs}\cdot\mathcal{L}_{vs},
\end{equation}where $\lambda_{a}^{ts}$ and $\lambda_{a}^{vs}$ are two trade-off hyper-parameters. With the additional alignment losses, the final objective function equals the sum of the cross-entropy loss with the alignment losses:

\section{Experiment}
In this section, we first introduce the experimental settings from four aspects, including datasets, evaluation metrics, implementation, and compared baselines. Then, we comprehensively analyze the proposed \sysname{} by answering the following questions. 
\begin{itemize}
    \item \textbf{Q1: Superiority.} Does \sysname{} outperform the existing state-of-the-art existing multimodal knowledge graph reasoning models, especially for the transformer models?
    \item \textbf{Q2: Effectiveness.} Are the adopted structure encoder and structure-guided fusion modules effective in fusing structure information into the MPT model for better MKGR performance? 
    \item \textbf{Q3: Efficiency.} Will the additional mechanism for structure information raise the unnecessary model and running time complexity for our \sysname{} compared to MKGformer?
    \item \textbf{Q4: Sensitivity.} How does the performance fluctuation of \sysname{} with different hyper-parameters?
\end{itemize}

We conduct experiments to answer the above questions. Specifically, answers of \textbf{Q1} to \textbf{Q4} are offered in Sec. 4.2 to 4.5.

\subsection{Experiment Setting}
\subsubsection{Datasets}
Two commonly used available datasets, \textit{i.e.,} FB15K-237-IMG and WN18-IMG, for multimodal knowledge graph relation reasoning are used in this paper. These datasets include three modalities, including text descriptions, corresponding images, and graph structures. Specifically, both FB15k-237-IMG \cite{MTRL} and WN18 \cite{MTRL} datasets are constructed by extending ten images for each entity based on FB15k-237 and WN18, which are the subset of the large-scale knowledge graph Freebase \cite{FreeBase} and WordNet \cite{WordNet}, separately. The detailed statistics of these two datasets are shown in Tab. \ref{ashit1}.

\subsubsection{Implementation Details}
The experiments are implemented on the computer with an Intel(R) Core(TM) i9-9900K CPU @ 3.60GHz, 64GB RAM, and one GeForce RTX 3090 Ti GPU using PyTorch 1.10.0 in CUDA 11.1. The model parameters are optimized using Adam \cite{Adam} optimizer, and we conduct a grid search to find suitable hyperparameters. We select MKGformer as our backbone. Following the MKGformer, we adopt the BERT base \cite{BERT} and ViT-B/32 \cite{VIT} as the text encoder and vision encoder in our method. Besides, a multimodal information encoder is chosen as the M-Encoder in \cite{MKGformer}. As for the structure encoder, most of our experiments are carried out based on the HAKE \cite{HAKE}, but we also evaluate the performance of our model with other typical structure encoders, including HousE \cite{li2022house}, and COMPGCN \cite{COMPGCN}. Besides, hyper-parameter $\lambda_{s}^{ts}$, $\lambda_{s}^{vs}$, $\lambda_{a}^{ts}$ and $\lambda_{a}^{vs}$ are set as 0.01, 0.01, 0.001, and 0.001. Following previous works \cite{MKGformer, RSME}, we use Hits@k, where k $\in$ \{1, 3, 10\}, and Mean Rank (MR) to evaluate our model, and the mean results of three runs of each experiment are reported.
\begin{itemize}
    \item Hits@k: Hits@k indicates the percentage of true entities that are included in the top k positions of a sorted rank list. 
    \begin{align}
    {Hits@k}={\frac{1}{|\mathcal{G}|}\sum_{r\in\mathcal{G}}^{}\mathbbm{I} [r\leq k]}
    \end{align}where $|\mathcal{G}|$ and $r_i$ represents the fact quantity and the rank for $i^{th}$ fact, separately. Besides, $\mathbbm{I}$ is an indicator function. Note that Hits@k $\in$ (0,1] does not consider the cases whose rank is larger than k, where closer to 1 is better. 
\end{itemize}
\begin{itemize}
    \item MR: Mean rank is the arithmetic mean over all individual ranks. 
    \begin{align}
    {MR}={\frac{1}{|\mathcal{G}|}\sum_{r\in\mathcal{G}}^{}r=\frac{1}{|\mathcal{G}|}(r_{1}+r_{2}+\cdots+r_{|\mathcal{G}|})}
    \end{align}
    Compared to Hits@k, MR $\in$ [1,+$\infty$) is more sensitive to any model performance changes, not only what occurs under a certain cutoff, where lower is better.
\end{itemize}

\begin{table}[!t]
\renewcommand\arraystretch{1.2}
    \caption{Statistics of FB15k-237-IMG and WN18-IMG}
    \vspace{-1em}
    \fontsize{6}{6.5}\selectfont 
    \resizebox{\linewidth}{!}{
\begin{tabular}{cccccc}
\toprule
Dataset       & \#Rel. & \#Ent. & \#Train & \#Dev & \#Test \\ \midrule
FB15k-237-IMG & 237    & 14541  & 272115  & 17535 & 20466  \\
WN18-IMG      & 18     & 40943  & 141442  & 5000  & 5000   \\ \bottomrule
\end{tabular}
}
\label{ashit1}
\end{table}
\begin{table*}[t]
\caption{Performance comparison of different KGR models for MKGR task on FB15K-237-IMG and WN18-IMG. The best results are in boldface and the second-best results are marked with the underline. Note that the Hit@k is presented in percentage.}
\vspace{-1 em}
\fontsize{4.5}{5.5}\selectfont 
 \label{performance}
\resizebox{\linewidth}{!}{
\begin{tabular}{ccccccccccc}
\hline
\multirow{2}{*}{Model} &  & \multicolumn{4}{c}{FB15k-237-IMG} &  & \multicolumn{4}{c}{WN18-IMG} \\ \cline{3-6} \cline{8-11} 
&  & MR & Hits@1   & Hits@3  & Hits@10     &  & MR & Hits@1 & Hits@3 & Hits@10  \\ \hline
\multicolumn{11}{c}{\textit{Non-Transformer KGR Models}}                                                \\ \hline
TransE                 &  & 323 & 19.8   & 37.6  & 44.1     &  & 357 & 4.0  & 74.5 & 92.3   \\
DisMult               &  & 512 & 19.9   & 30.1  & 44.6     &  & 665 & 33.5 & 87.6 & 94.0    \\
ComplEx            & & 546 & 19.4   & 29.7  & 45.0      &  & - & 93.6 & 94.5 & 94.7    \\ 
ConvE             &  & 249 & 22.5   & 34.1  & 49.7     &  & - & 41.9 & 47.0 & 53.1    \\
RGCN                &  & 600 & 10.0   & 18.1  & 30.0     &  & - & 8.0 & 13.7 & 20.7    \\
IKRL(UNION)            &  & 298 & 19.4   & 28.4  & 45.8     &  & 596 & 12.7 & 79.6 & 92.8   \\
TransAE                & & 431 & 19.9   & 31.7  & 46.3     &   & 352& 32.3 & 83.5 & 93.4  \\
RSME(ViT-B/32+Forget)  &  & 417 & 24.2   & 34.4  & 46.7     &   &-& \bf{94.3} & 95.1 & -    \\ \hline
\multicolumn{11}{c}{\textit{Transformer KGR Models}}                                       \\ \hline
KG-BERT                &  & \bf{153} & -       & -      & 42.0      &   & 58 & 11.7 & 68.9 & 92.6   \\
VisualBERT             &  & 592 & 21.7   & 32.4  & 43.9     &  & 122 & 17.9 & 43.7 & 65.4   \\
ViLBERT                &  & 483 & 23.3   & 33.5  & 45.7     &  & 131  & 22.3 & 55.2 & 76.1  \\
MKGformer              &  & 252 & \underline{24.3}   & \underline{36.0}   & \underline{49.9}     &   & \bf{25}  & \underline{93.5} & \underline{95.8} & \underline{97.0}  \\
\sysname{} (Ours)               &  & \underline{238} & \bf{25.2}   & \bf{37.0}  & \bf{51.0}     &  & \underline{29}  & \bf{94.3} & \bf{96.6} & \bf{97.8}    \\ \hline
\end{tabular}
}
\end{table*}

\subsubsection{Compared Baselines}
The compared models include two types, \textit{i.e.,} non-transformer KGR models and transformer KGR models. 
Among the models in the first type, there are five unimodal KGR state-of-the-art models, including TransE \cite{TransE}, DisMult \cite{DisMult}, ComplEX \cite{ComplEX}, ConvE \cite{ConvE}, and RGCN \cite{RGCN}, and three multimodal state-of-the-art KGR models, including IKRL \cite{IKRL}, TransAE \cite{TransAE} and RSME \cite{RSME}. As for the transformer models, there are three multimodal models except the KG-BERT \cite{KG-BERT}, which is the first unimodal KGR model developed based on transformer architecture. VisualBERT \cite{visualbert} and ViLBERT \cite{Vilbert} are the general multimodal models that can also be applied to multimodal KGR. Besides, MKGformer is the representative multimodal KGR model with transformer architecture. However, none of the transformer KGR models makes use of the structure information except for our \sysname{}.
\begin{table}[t]
\caption{Ablation study of \sysname{} on FB15K-237-IMG. 'WS' and 'AC' represent the weighted summation and alignment constraint. The Hit@k is presented in percentage.}
\fontsize{10}{13}\selectfont 
 \label{ablation}
\resizebox{\linewidth}{!}{
\begin{tabular}{lcccc}
\hline
\multirow{2}{*}{Model}    & \multicolumn{4}{c}{FB15k-237-IMG} \\ \cline{2-5} 
& MR    & Hits@1  & Hits@3  & Hits@10  \\ \hline
SGMPT        & 238   & 25.2   & 37.0     & 51.0      \\
\rowcolor{lime} - WS$^{ts}$   & 242   & 24.7   & 36.6    & 50.6    \\
\rowcolor{lime} - WS$^{vs}$   & 240   & 25.0  & 36.8  & 50.8    \\
\rowcolor{pink} - AC$^{ts}$   & 242   & 24.7   & 36.6   & 50.6      \\
\rowcolor{pink} - AC$^{vs}$   & 241   & 24.8   & 36.8   & 50.7  \\
\rowcolor{lime} - (WS$^{ts}$ \& WS$^{vs}$)     & 247   & 24.5   & 36.3   & 50.4    \\
\rowcolor{pink} - (AC$^{ts}$ \& AC$^{vs}$)    & 248   & 24.4   & 36.2  & 50.2    \\
\rowcolor{yellow} - (WS$^{ts}$ \& AC$^{ts}$)   & 248   & 24.4   & 36.2   & 50.2    \\
\rowcolor{yellow} - (WS$^{vs}$ \& AC$^{vs}$)    & 245   & 24.6   & 36.5   & 50.5    \\
\rowcolor{lightgray} - (WS$^{ts}$ \& WS$^{vs}$ \& AC$^{ts}$ \& AC$^{vs}$) & 252   & 24.3   & 36.0     & 49.9    \\ \hline
\end{tabular}}
\end{table}

\subsection{Performacne Comparsion (RQ1)}
The overall performance comparison is carried out to answer \textbf{Q1}. We compare our \sysname{}  with thirteen other state-of-the-art models on two benchmark datasets. The results in Tab.\ref{performance} show that our \sysname{} outperforms all of the evaluation metrics compared to non-transformer KGR models. Besides, compared to the transformer KGR models, our \sysname{} outperforms most of the evaluation metrics compared to transformer KGR models. In particular, it is more apparent that our model boosts the performance on FB15k-237-IMG. Concretely, our model makes 4.1\%, 7.6\%, and 2.6\% improvements on Hits@1, Hits@3, and Hits@10 compared to non-transformer KGR models, and 3.7\%, 2.8\%, and 2.2\% improvements on Hit@1, Hits@3, and Hits@10 compared to transformer KGR models, respectively. It shows the superiority of our model \sysname{}. Besides, it also indicates the advances of the transformer paradigm for the MKGR task, since the average performances of transformer KGR models are better than non-transformer KGR models in the multimodal scenario. Although the MR values of our model are not the best, they still occupy the second-best positions. Moreover, our MR performances are actually better and comparable compared to the backbone MPT model, \textit{i.e.,} MKGformer. The above results show that our \sysname{} can achieve better reasoning performance in the multimodal scenario, which further indicates the effectiveness of involving the structural information in MPT models.
\begin{figure}[t]
\centering
\includegraphics[width=0.48\textwidth]{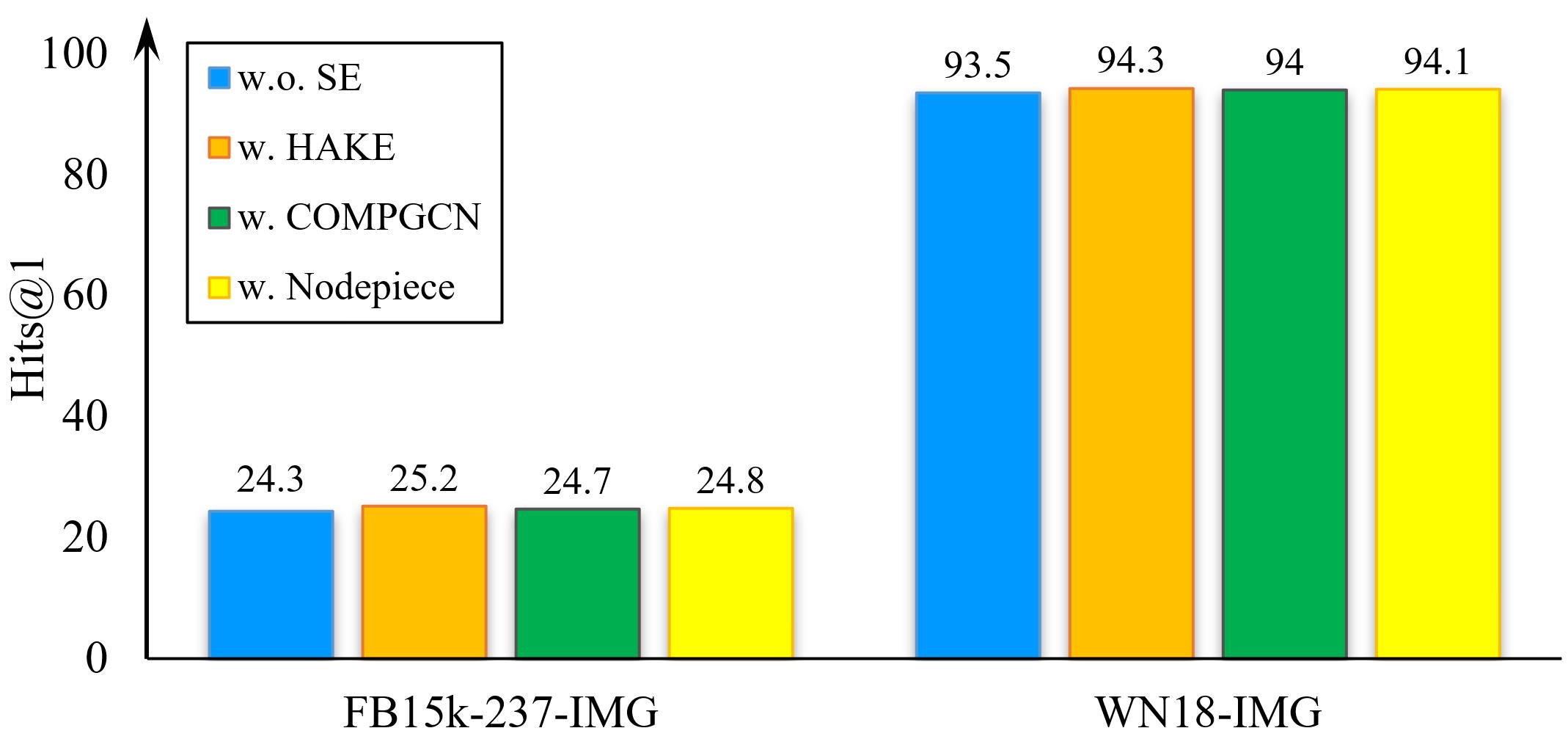}
\caption{Performance comparison with different structure encoders. 'w.' and 'w.o.' represent the with and without separately. 'SE' represents the structure encoder.}
\label{dsf}  
\end{figure}

\subsection{Ablation Study (RQ2)}
The ablation studies are conducted on FB15k-237-IMG to answer the Q2. Specifically, two subquestions need to be answered in the following subsections, \textit{i.e.,} \textbf{(1)} "Can the structure-guided fusion modules make differences?" and \textbf{(2)} "Will different structure encoders benefit the structure information fusion procedure?".

\subsubsection{Effectiveness of the Structure-guided Fusion Module}
We evaluate the effectiveness of the structure-guided fusion module from four parts, \textit{i.e.,} weighted summation, alignment constraint, text-structure fusion, and vision-structure fusion. Concretely, ten sub-models are compared, including (1) the original \sysname{} model, (2) \sysname{} without text-structure weight summation denoted as "- WS$_{ts}$", (3) \sysname{} without vision-structure weight summation denoted as "- WS$_{vs}$", (4) \sysname{} without text-structure alignment constraint denoted as "- AC$_{vs}$", (5) \sysname{} without vision-structure alignment constraint denoted as "- AC$_{vs}$", (6) \sysname{} without weight summation denoted as "- (WS$^{ts}$ \& WS$^{vs}$)", (7) \sysname{} without alignment constraint summation denoted as "- (AC$^{ts}$ \& AC$^{vs}$)", (8) \sysname{} without text-structure strategies denoted as "- (WS$^{ts}$ \& AC$^{ts}$)", (9) \sysname{} without vision-structure strategies denoted as "- (WS$^{vs}$ \& AC$^{vs}$)", (10) \sysname{} without structure-guided fusion module denoted as "- (WS$^{ts}$ \& WS$^{vs}$ \& AC$^{ts}$ \& AC$^{vs}$)". Tab. \ref{ablation} shows that performance boosts are made by the strategies used in the structure-guided fusion module, \textit{i.e.,},  5.9\%, 3.7\%, 2.8\%, and 2.2\% improvements for MR, Hits@1, Hits@3, and Hits@10. More specifically, the alignment constraint and weighted summation are relatively equivalently effective to MKGR. Besides, vision-structure fusion is less than text-structure fusion, due to the information loss caused by the vector expanding operations. In all, the promising results demonstrate the effectiveness of the module and also prove that strategies can be adopted either individually or composedly.

\begin{table}[t]
\caption{Number of Model parameters and running time comparison between our \sysname{} and backbone MKGformer. Note that H represents hour.}
\fontsize{9}{13}\selectfont 
 \label{ablation}
\resizebox{\linewidth}{!}{
\begin{tabular}{cclll}
\hline
\multirow{2}{*}{Model} & \multicolumn{2}{c}{FB15k-237-IMG}                                & \multicolumn{2}{c}{WN18-IMG} \\ \cline{2-5} 
& \multicolumn{1}{l}{\# Param.} & Time                  & \# Param.  & Time \\ \hline
MKGformer              &   950.540 M                      &   9.6 H                &       1029.818 M     & 24.3 H         \\
SGMPT                  & 956.687 M                         & \multicolumn{1}{c}{9.8 H} &    1032.893 M     & 25.4 H             \\ \hline
\end{tabular}}
\end{table}

\subsubsection{Influence of Different Structure Encoder}
We also replace the structure encoder $g(\cdot)$, \textit{i.e.,} HAKE \cite{HAKE}, as three other typical structure encoders, including CompGCN \cite{COMPGCN}, and Nodepiece \cite{galkin2022nodepiece}. Experiments are further conducted based on FB15k-237-IMG for Hit@1 and Hit@10 (See Fig. \ref{dsf}). It indicates that various structure encoders can all benefit the structure information fusion procedure, \textit{i.e.,} on average 2.5\% and 2.0\% improvements on FB15k-237-IMG and WN18-IMG separately. Besides, the HAKE is the most effective choice among these three structure encoders.

Based on promising results and the above analyses in Sec 4.3.1 and Sec. 4.3.2, we can assert that both the adopted structure encoder and designed structure-guided fusion modules are effective in fusing the omitted structure information in KG into the MPT model for better MKGR performance.

\subsection{Efficiency Analysis (RQ3)}
We analyze the complexity of our \sysname{} from two aspects, \textit{i.e.,} parameter number and the running time of the models. Concretely, the comparison is carried out between our \sysname{} and the backbone MKGformer on both FB15k-237-IMG and WN18-IMG. According to Tab. \ref{dsf}, it is observed that the efficiency of our model is a little bit worse than MKGformer, \textit{i.e.,} average 0.45\% and 3.3\% increasing on the number of parameters and running time, respectively. Since the structure fusion mechanism is designed to complement the omitted structure information, our \sysname{} will definitely be more complex and time-consuming. However, considering the performance improvements for MKGR, it is acceptable with the comparable complexity, which indicates that our \sysname{} does not raise unnecessary parameters and running time. 
\begin{figure}[t]
\centering
\begin{minipage}{\linewidth}
\centerline{\includegraphics[width=\textwidth]{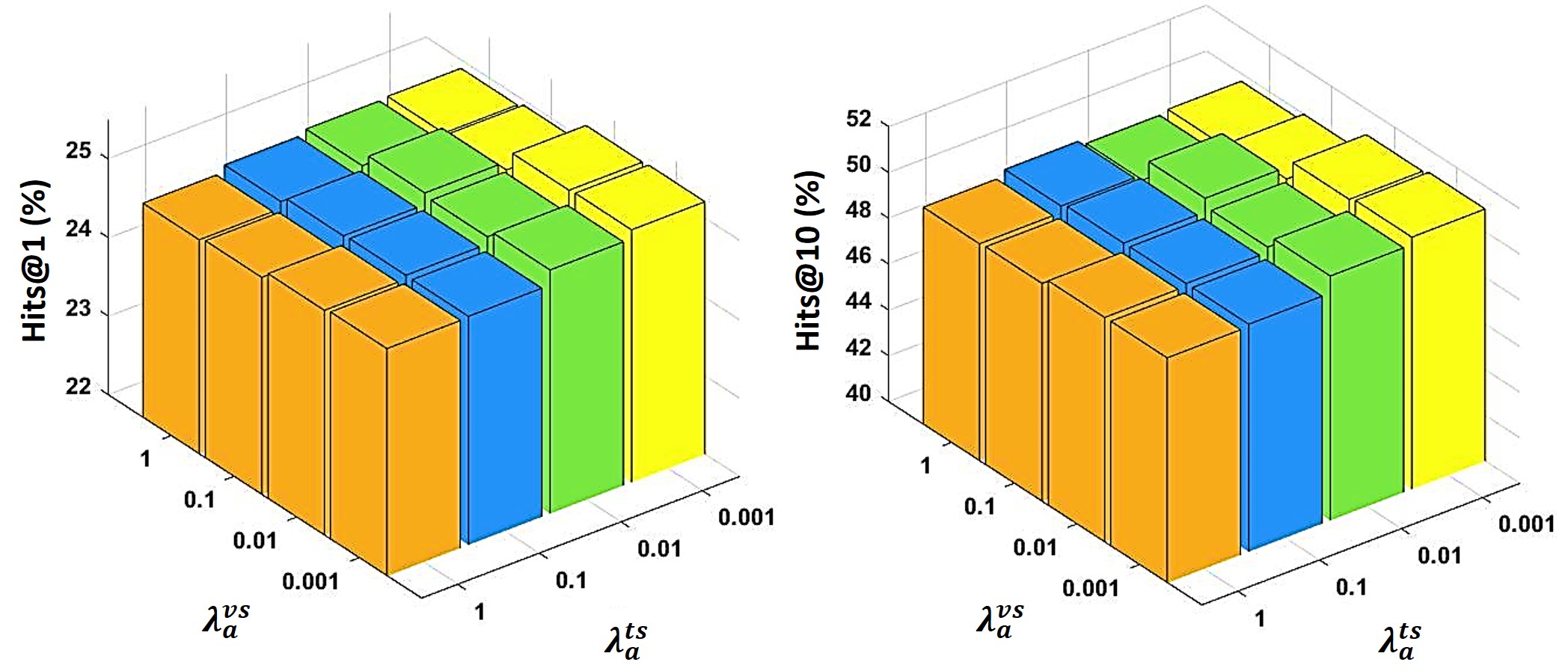}}
\vspace{1pt}
\centerline{(a) hyper-parameters analysis for weighted summation.}
\end{minipage}
\begin{minipage}{\linewidth}
\centerline{\includegraphics[width=\textwidth]{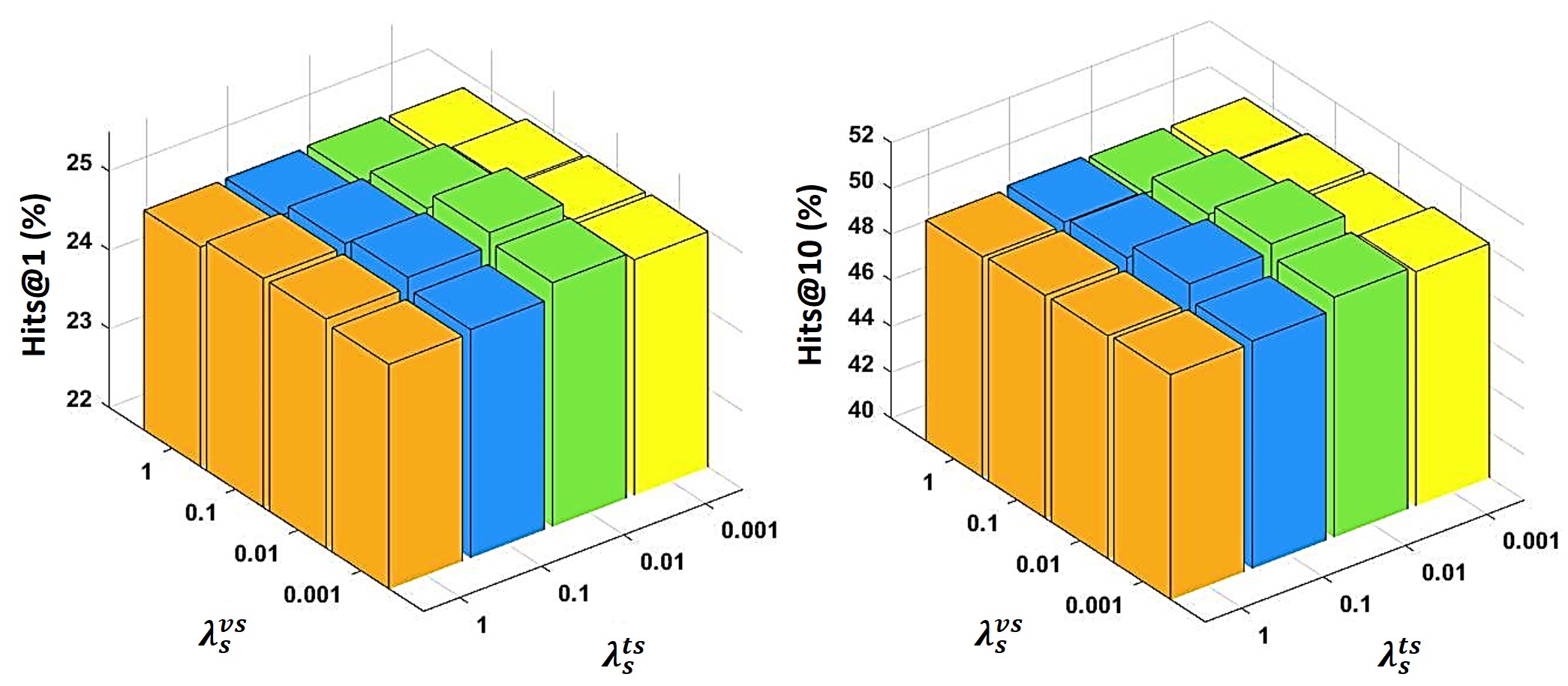}}
\vspace{1pt}
\centerline{(b) hyper-parameters analysis for alignment constraint.}
\end{minipage}
\caption{Sensitivity analysis of hyper-parameter $\lambda_{s}^{ts}$, $\lambda_{s}^{vs}$, $\lambda_{a}^{ts}$, $\lambda_{a}^{vs}$ on FB15k-237-IMG for Hit@1 and Hit@10. Note that $\lambda_{a}^{ts}$, $\lambda_{a}^{vs}$ both equal to 0.001 in subgraph (a) and $\lambda_{s}^{ts}$, $\lambda_{s}^{vs}$ both equal to 0.01 in subgraph (b).}
\vspace{-1 em}
\label{beta}
\end{figure}

\subsection{Sensitivity Analysis (RQ4)}
We investigate the influence of the hyper-parameter $\lambda_{s}^{ts}$, $\lambda_{s}^{vs}$, $\lambda_{a}^{ts}$, $\lambda_{a}^{vs}$ on FB15k-237-IMG for Hit@1 and Hit@10. Besides, the scope of these four hyper-parameters is selected in $\{0.001, 0.01, 0.1, 1 \}$. We observe that the MKGR performance will not fluctuate greatly when the parameter is varying in Fig. \ref{beta}. It demonstrates that the performance of \sysname{} is insensitive to these hyper-parameters. We can further find out that the best performance is reached with the combination of $\lambda_{s}^{ts}$, $\lambda_{s}^{vs}$, $\lambda_{a}^{ts}$ and $\lambda_{a}^{vs}$ set to (0.01, 0.01, 0.001, 0.001).

\section{Conclusion}
In this paper, we propose a novel and simple graph \underline{S}tructure \underline{G}uided \underline{M}ultimodal \underline{P}retrained \underline{T}ransformer model for knowledge graph reasoning, termed \sysname{}. As the first multimodal pretrained transformer model to leverage the structure information for multimodal knowledge graph reasoning, \sysname{} is the first work to add a specific structure information fusion procedure based on the MPT backbone. To complete the procedure, the structure encoder and structure-guided fusion module are required. More concretely, various KGE models can be selected as the structure encoder, and two different strategies, \textit{i.e.,} weighted summation and alignment constraint, can be adopted both individually and composedly. Extensive experiments on two benchmark datasets, \textit{i.e.,} FB15k-IMG and WN18-IMG, demonstrate the promising capacity of our \sysname{} from four aspects, including superiority, effectiveness, complexity, and sensitivity. In the future, we plan to continue to improve the capacity of our model toward its limitations. For example, four hyperparameters are enrolled to complement the omitted structural information. Moreover, the weighted summation strategy can be more fine-grained though the current simple addition can also achieve promising improvements.

\bibliographystyle{ACM-Reference-Format}
\bibliography{sample-base}
\end{document}